\documentclass[twoside,11pt]{article}

\usepackage{jmlr2e}
\usepackage{graphicx}
\usepackage{float}
\usepackage{listings}
\usepackage{mathtools}
\usepackage{pgfplots}
\usepackage[super]{nth}

\pgfkeys{/pgf/declare function={arcsinh(\x) = ln(\x + sqrt(\x^2+1));}}
\pgfplotsset{compat=1.14}

\ShortHeadings{hyper-sinh: An Accurate and Reliable Function from Shallow to Deep Learning}{Luca Parisi, Renfei Ma, Narrendar RaviChandran, Matteo Lanzillotta}
\firstpageno{1}

\begin{document}

\title{hyper-sinh: An Accurate and Reliable Function from Shallow to Deep Learning in TensorFlow and Keras}

\author{\name Luca Parisi, PhD, MBA Candidate \email luca.parisi@ieee.org \\
       \addr Faculty of Business and Law (Artificial Intelligence Specialism) \\
       Coventry University \\
       Coventry, United Kingdom \\
       \\
       \addr University of Auckland Rehabilitative Technologies Association (UARTA) \\
       University of Auckland \\
       11 Symonds Street, Auckland, 1010, New Zealand \\
       \AND
       \\
       \name Renfei Ma, PhD \email marenfei@cuhk.edu.cn \\
       \addr Warshel Institute for Computational Biology\\
       The Chinese University of Hong Kong, Shenzhen (CUHK-SZ)\\
       Shenzhen, China \\
       \\
       \addr University of Auckland Rehabilitative Technologies Association (UARTA) \\
       University of Auckland \\
       11 Symonds Street, Auckland, 1010, New Zealand
       \AND
       \\
       \name Narrendar RaviChandran, PhD \email narrendar@ieee.org \\
       \addr University of Auckland Rehabilitative Technologies Association (UARTA) \\
       University of Auckland \\
       11 Symonds Street, Auckland, 1010, New Zealand
       \AND
       \\
       \name Matteo Lanzillotta, MSc \email amatt.do@gmail.com \\
       \addr Department of Counselling Psychology and Psychotherapy \\
       Centro Studi Eteropoiesi \\
       Turin, Italy \\
       \\
       \addr University of Auckland Rehabilitative Technologies Association (UARTA) \\
       University of Auckland \\
       11 Symonds Street, Auckland, 1010, New Zealand}

\editor{}

\maketitle
\newpage

\begin{abstract}
This paper presents the 'hyper-sinh', a variation of the m-arcsinh activation function suitable for Deep Learning (DL)-based algorithms for supervised learning, such as Convolutional Neural Networks (CNN).
hyper-sinh, developed in the open source Python libraries \emph{TensorFlow} and \emph{Keras}, is thus described and validated as an accurate and reliable activation function for both shallow and deep neural networks. 
Improvements in accuracy and reliability in image and text classification tasks on five (N = 5) benchmark data sets available from Keras are discussed. 
Experimental results demonstrate the overall competitive classification performance of both shallow and deep neural networks, obtained via this novel function. 
This function is evaluated with respect to gold standard activation functions, demonstrating its overall competitive accuracy and reliability for both image and text classification.
\end{abstract}

\vskip 0.1in

\begin{keywords}
  Activation, Deep Learning, Convolutional Neural Network, TensorFlow, Keras
\end{keywords}


\vskip 0.3in

\section{Introduction}

\vskip 0.15in

Despite recent developments of activation functions for Machine Learning (ML)-based classifiers, such as the m-arcsinh~\citep{parisi2020m} for shallow Multi-Layer Perceptron (MLP)~\citep{rumelhart1986learning}, usable, repeatable and reproducible functions for both shallow and deep neural networks, e.g., the Convolutional Neural Network (CNN)~\citep{lecun1995convolutional}, have remained very limited and confined to three activation functions regarded as 'gold standard'. These include the Rectified Linear Unit (ReLU), the sigmoid function and its modified version, hyperbolic tangent sigmoid or 'tanh'~\citep{lin2003study}, which extends its range from [0, +1] to [-1, +1]. The sigmoid and tanh have well-known vanishing gradient issues; thus, the ReLU function was devised to be more scalable for deep neural networks, despite its 'dying ReLU' problem, which has recently been solved by~\citep{parisi2020qrelu}. These have been made freely accessible in the open source Python library named 'Keras'~\citep{chollet2015keras} for Deep Learning. The availability of these functions in the public domain has enabled not-for-profit and for-profit organisations to leverage them for several intelligence-based applications, from academic to industrial applications~\citep{chollet2017xception}~\citep{parisi2020qrelu}.
\\

Nevertheless, considering the above-mentioned challenges in the Computer Science and ML communities, such activation functions lack robustness with classification tasks of varying degrees of complexity, e.g., slow or lack of convergence~\citep{vert2006consistency} ~\citep{jacot2018neural}, caused by trapping at local \emph{minima}~\citep{parisi2020novel}. Moreover, amongst the three above-mentioned activation functions, only the ReLU is applicable from shallow to deep neural networks, with its novel quantum variations (QReLU and m-QReLU) found more scalable than its traditional version only recently~\citep{parisi2020qrelu}.
\\

On the other side, in sciences dealing with the study of human behaviour, in the last 20 years, considerable progress has been made towards the prevention of mental health disorders~\citep{sander2016effectiveness}~\citep{ebert2017prevention}.
Specifically, professionals working in the field of counselling psychology have slightly enhanced their ability of grasping relational issues in their subjects via novel ML-based tele-monitoring technologies~\citep{shatte2019machine}.
Nevertheless, these technologies have not yet changed the traditional counselling psychology practice, which is still based on a structured methodology that is adopted to help individuals to become more self-aware, more conscious of their own needs and moods~\citep{pieterse2013towards}. The main goal counsellors pursue is guiding individuals to get to know themselves at a deeper level and to help them discover and resurface their own resources to better manage their emotions in their daily life. This process first requires a tailored dialogue between the counsellor and the individual and, subsequently, leveraging practical tools to aid the individual in their experience to understand their inner self more deeply~\citep{sutton2016measuring}.
Moreover, there are still limitations within the counselling setting. For instance, individuals, out of fear, may not reveal fundamental aspects of their \emph{persona} that would help counsellors guide them better in getting to know themselves. Furthermore, in many cases, subjects may express a verbal language opposite to their non-verbal one. Counsellors often hardly understand the dynamic patterns observed in the behaviours of their subjects, thus being unable to provide the required help and support to them.
\\

In counselling, neural network algorithms, both shallow and deep depending on the amount of good-quality data and hardware available, have the potential to support counsellors in image and text classification tasks to understand and guide their subjects by helping them infer subtle dynamic changes in their behaviours. Via a careful and effective observation of images, micro- and macro- body movements, and facial expressions~\citep{lee2017chatbot}~\citep{oh2017chatbot}, it is possible to better interpret and understand the subjects' non-verbal language. Even the emotions underlying the written content from subjects may reveal inner aspects of their \emph{persona} that are fundamental for counsellors to help resurface to increase the subjects' self-awareness and related capability of 'self-healing'~\citep{rennie2001client}.
\\

Therefore, from both theoretical and practical standpoints, there is an increasing need for accurate and reliable open source activation functions, which reach convergence faster, avoiding trapping at local \emph{minima}, are more stable and can also be used and scale across both shallow and deep neural network algorithms for image and text classification. Entirely written in Python and made freely available in TensorFlow~\citep{abadi2016tensorflow} and Keras~\citep{chollet2015keras}, the proposed hyperbolic function is demonstrated as a competitive function with respect to gold standard functions, which suits both shallow and deep neural networks, thus being accurate and reliable for pattern recognition to aid image and text classification tasks.

Thanks to its liberal license, it has been widely distributed as a part of the free software Python libraries TensorFlow~\citep{abadi2016tensorflow} and Keras~\citep{chollet2015keras}, and it is available for use for both academic research and commercial purposes.\\


\vskip 0.3in

\section{Methods}

\vskip 0.15in

\subsection{Data sets used from Keras}

\vskip 0.15in

The following benchmark data sets for image and text classification from \emph{Keras}~\citep{chollet2015keras} were used in the experiments described and discussed in this study:

\begin{itemize}

\item 'CIFAR-10' data set~\citep{krizhevsky2009learning}, having 50,000 32x32 colour images for training, and 10,000 images for testing, labelled based on 10 mutually exclusive classes of corresponding objects, including airplanes, automobiles, birds, cats, deers, dogs (e.g., sedans, SUVs, etc.), frogs, horses, ships, trucks (only big trucks);
\item 'Fashion-MNIST' data set~\citep{xiao2017fashion}, which has 60,000 28x28 grayscale images of 10 classes of fashion (T-shirts/tops, trousers, pullovers, dresses, coats, sandals, shirts, sneakers, bags, ankle boots), with 10,000 images for testing;
\item 'MNIST' data set~\citep{lecun1998mnist}, with 60,000 28x28 grayscale images of the 10 handwritten digits, having 10,000 images for testing;
\item 'Reuters' data set~\citep{apte1994automated}, which has 11,228 news-wires from Reuters, labelled over 46 classes of topics. Each news-wire is encoded as a list of word indices based on their overall frequency in the data set. '0' (zero) is used to encode any unknown words. Words not seen in the training set but that are present in the test set have been skipped.
\item 'IMDB' data set~\citep{maas2011learning}, with 25,000 pre-processed movies reviews from IMDB, labelled by sentiment (positive or negative). Each review is encoded as a list of word indexes (integers) based on their overall frequency in the data set. '0' (zero) is used to encode any unknown words.
 
\end{itemize}

\vskip 0.15in

\subsection{Baseline neural network models and hyperparameters}

\vskip 0.15in

As the purpose of this study is not to devise the most optimised, best-performing classifier for any of the classification tasks involved in sub-section 2.1, but, instead, to extend the m-arcsinh into a novel accurate and reliable activation function that can scale from shallow to deep neural networks, and evaluate it against the current gold standard functions available in the Python library \emph{Keras}~\citep{chollet2015keras}, baseline Fully-Connected Neural Networks (FC-NN) and Convolutional Neural Networks (CNN) models were used with the following hyperparameters for the respective classification tasks in sub-section 2.1. The activation functions in the convolutional layers were made vary for testing purposes across the following: ReLU, sigmoid, tanh and the proposed hyper-sinh.

The CNN-related hyperparameters to classify the CIFAR-10 data set are as follows:

\begin{itemize}
\item three convolutional layers, each of which has a kernel size of 3x3;
\item the following convolutional filters for each of the three convolutional layers (in order from the first layer to the third one): 32, 64, 64;
\item max pooling is applied after the first and the second convolutional layers;
\item after a flattening layer, two dense layers follow, the first one having 64 neurons and ReLU activation, the second one having 10 neurons as per the number of classes in the CIFAR-10 data set.

\end{itemize}

Listing 1 provides the snippet of code in Python to use a CNN to classify the CIFAR-10 data set, with different activation functions available in \emph{Keras}~\citep{chollet2015keras}, including the novel 'hyper-sinh'.

\lstset{language=Python}
\lstset{frame=lines}
\lstset{caption={Python code to use a CNN to classify the CIFAR-10 data set, with different activation functions available in \emph{Keras}~\citep{chollet2015keras}, including the proposed 'hyper-sinh'.}}
\lstset{label={lst:code_direct_1}}
\lstset{basicstyle=\footnotesize}
\begin{lstlisting}

from tensorflow.keras import models, layers

model = models.Sequential()

# First convolutional layer with ReLU, sigmoid, or tanh activation function

model.add(layers.Conv2D(32, (3, 3), activation='relu', input_shape=(32, 32, 3)))

# First convolutional layer with custom layer, if hyper-sinh were used as 
# activation function

# model.add(layers.Conv2D(32, (3, 3), input_shape=(32, 32, 3)))
# model.add(hyper_sinh())

model.add(layers.MaxPooling2D((2, 2)))

# Second convolutional layer with ReLU, sigmoid, or tanh activation function

model.add(layers.Conv2D(64, (3, 3), activation='relu'))

# Second convolutional layer with custom layer, if hyper-sinh were used as 
# activation function

# model.add(layers.Conv2D(64, (3, 3)))
# model.add(hyper_sinh())

model.add(layers.MaxPooling2D((2, 2)))

# Third convolutional layer with ReLU, sigmoid, or tanh activation function

model.add(layers.Conv2D(64, (3, 3), activation='relu'))

# Third convolutional layer with custom layer, if hyper-sinh were used as 
# activation function

# model.add(layers.Conv2D(64, (3, 3)))
# model.add(hyper_sinh())

# Flattening and dense layers, with the last one for classification having 10
# neurons as per the number of classes of the CIFAR-10 data set
model.add(layers.Flatten())
model.add(layers.Dense(64, activation='relu'))
model.add(layers.Dense(10))

\end{lstlisting}

\vskip 0.15in

The FC-NN-related hyperparameters to classify the Fashion-MNIST data set are as follows:

\begin{itemize}
\item one flattening layer;
\item one dense layer with 128 neurons, with varying activation based on the testing case scenario (one amongst sigmoid, tanh, ReLU and the proposed hyper-sinh);
\item a final dense layer having 10 neurons as per the number of classes in the Fashion-MNIST data set.

\end{itemize}

Listing 2 provides the snippet of code in Python to use a FC-NN to classify the Fashion-MNIST data set, with different activation functions available in \emph{Keras}~\citep{chollet2015keras}, including the novel 'hyper-sinh'.

\lstset{language=Python}
\lstset{frame=lines}
\lstset{caption={Python code to use a FC-NN to classify the Fashion-MNIST data set, with different activation functions available in \emph{Keras}~\citep{chollet2015keras}, including the proposed 'hyper-sinh'.}}
\lstset{label={lst:code_direct_2}}
\lstset{basicstyle=\footnotesize}
\begin{lstlisting}

from tensorflow.keras import models, layers

model = models.Sequential()

model.add(layers.Flatten(input_shape=(28, 28)))

# First dense layer with ReLU, sigmoid, or tanh activation function

model.add(layers.Dense(128, activation='relu'))

# First dense layer with custom layer, if hyper-sinh were used as 
# activation function

# model.add(layers.Dense(128))
# model.add(hyper_sinh())

# Second and final dense layer, with 10
# neurons for classification as per the number 
# of classes of the Fashion-MNIST data set

model.add(layers.Dense(10))

\end{lstlisting}

\vskip 0.15in

The CNN-related hyperparameters to classify the MNIST data set are as follows:

\begin{itemize}
\item two convolutional layers, each of which has a kernel size of 3x3;
\item the following convolutional filters for each of the two convolutional layers respectively (in order from the first layer to the second one): 32, 64;
\item max pooling is applied after the first and the second convolutional layers;
\item after a flattening layer, a dropout layer is leveraged with 0.5 (50\%) as dropout rate;
\item a final dense layer with softmax activation, having 10 neurons as per the number of classes in the MNIST data set.

\end{itemize}

Listing 3 provides the snippet of code in Python to use a CNN to classify the MNIST data set, with different activation functions available in \emph{Keras}~\citep{chollet2015keras}, including the novel 'hyper-sinh'.

\lstset{language=Python}
\lstset{frame=lines}
\lstset{caption={Python code to use a CNN to classify the MNIST data set, with different activation functions available in \emph{Keras}~\citep{chollet2015keras}, including the proposed 'hyper-sinh'.}}
\lstset{label={lst:code_direct_3}}
\lstset{basicstyle=\footnotesize}
\begin{lstlisting}

from tensorflow.keras import models, layers

model = models.Sequential()

# First convolutional layer with ReLU, sigmoid, or tanh activation function

model.add(layers.Conv2D(32, (3, 3), activation='relu', input_shape=(28, 28, 1)))

# First convolutional layer with custom layer, if hyper-sinh were used as 
# activation function

# model.add(layers.Conv2D(32, (3, 3), input_shape=(32, 32, 3)))
# model.add(hyper_sinh())

model.add(layers.MaxPooling2D((2, 2)))

# Second convolutional layer with ReLU, sigmoid, or tanh activation function

model.add(layers.Conv2D(64, (3, 3), activation='relu'))

# Second convolutional layer with custom layer, if hyper-sinh were used as 
# activation function

# model.add(layers.Conv2D(64, (3, 3)))
# model.add(hyper_sinh())

model.add(layers.MaxPooling2D((2, 2)))

# Flattening, dropout layer with 0.5 dropout rate, and 
# final dense layer for classification having 10 neurons 
# as per the number of classes of the MNIST data set
model.add(layers.Flatten())
model.add(layers.Dropout(0.5))
model.add(layers.Dense(10, activation="softmax"))

\end{lstlisting}

\vskip 0.15in

The FC-NN-related hyperparameters to classify the Reuters news-wires data set are as follows:

\begin{itemize}
\item one dense layer with 512 neurons, with varying activation based on the testing case scenario (one amongst sigmoid, tanh, ReLU and the proposed hyper-sinh);
\item a dropout layer is leveraged with 0.5 (50\%) as dropout rate;
\item a final dense layer with softmax activation, having 46 neurons as per the number of classes/topics in the Reuters news-wires data set.

\end{itemize}

Listing 4 provides the snippet of code in Python to use a FC-NN to classify the Reuters news-wires and the IMDB data sets, with different activation functions available in \emph{Keras}~\citep{chollet2015keras}, including the novel 'hyper-sinh'.

\lstset{language=Python}
\lstset{frame=lines}
\lstset{caption={Python code to use a FC-NN to classify the Reuters news-wires data set, with different activation functions available in \emph{Keras}~\citep{chollet2015keras}, including the proposed 'hyper-sinh'.}}
\lstset{label={lst:code_direct_4}}
\lstset{basicstyle=\footnotesize}
\begin{lstlisting}

from tensorflow.keras import models

model = models.Sequential()

# First dense layer having 512 neurons, with ReLU, sigmoid, tanh, or hyper-sinh 
# activation
model.add(Dense(512, input_shape=(10,000,)))
model.add(Activation('relu'))
# model.add(hyper_sinh())

# Dropout layer with 0.5 as dropout rate
model.add(Dropout(0.5))

# Second and final dense layer with softmax activation, 
# having 46 or 2 neurons for classification as per the number 
# of classes ('num_classes') of the Reuters news-wires or IMDB data sets 
# respectively
model.add(Dense(num_classes))
model.add(Activation('softmax'))

\end{lstlisting}

\vskip 0.15in

\newpage

\subsection{hyper-sinh: A reliable activation function for both shallow and deep learning}

For a function to be generalised as an activation function for both shallow and deep neural networks, such as FC-NN and CNN respectively, it has to be able to 1) avoid common gradient-related issues, such as the vanishing and exploding gradient problems and 2) improve discrimination of input data into target classes via a transfer mechanism of appropriate non-linearity and extended range. Considering the two-fold value of m-arcsinh~\citep{parisi2020m} as a kernel and activation function concurrently for optimal separating hyperplane- and shallow neural network-based classifiers, it was leveraged as the baseline function to be extended for it to scale to deep neural networks. Thus, although the arcsinh was swapped with its original sinh version, and the square root function was replaced with the basic cubic function, their weights were kept as per the m-arcsinh~\citep{parisi2020m} equivalent implementation, i.e., whilst 1/3 now multiplies sinh, 1/4 is now multiplying the cubic function.

Thus, the novel function \emph{hyper-sinh} was devised to be suitable for both shallow and deep neural networks concurrently by leveraging a weighted interaction effect between the hyperbolic nature of the hyperbolic sine function ('sinh') for positive values and the non-linear characteristic of the cubic function for negative values and 0 (zero), more suitable for deep neural networks, whilst retaining their appropriateness for shallow learning too, thus satisfying both the above-mentioned requirements:

\vspace{1em}
\begin{math}
x = \sinh{x} \times \frac{1}{3}, if x > 0 \hspace{26.5em}(1)
\end{math}
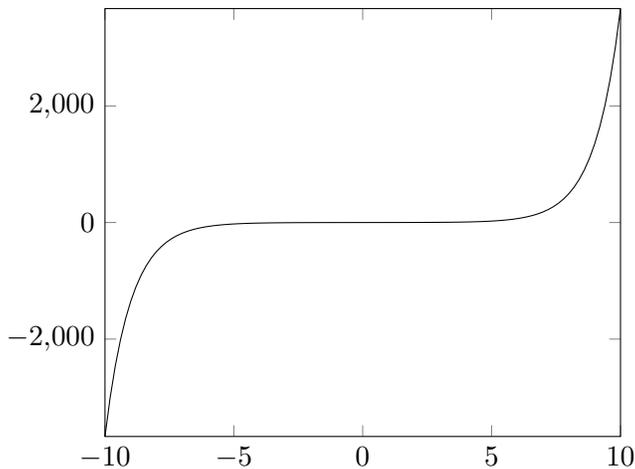
\vspace{2em}

\begin{tikzpicture}
\begin{axis}[enlargelimits=false]
\addplot [domain=-10:10, samples=101,unbounded coords=jump]{(sinh(x)/3)};
\end{axis}
\end{tikzpicture}

\vspace{2em}

\vspace{1em}
\begin{math}
x = x^3 \times \frac{1}{4}, if x \le 0 \hspace{28.5em}(2)
\end{math}
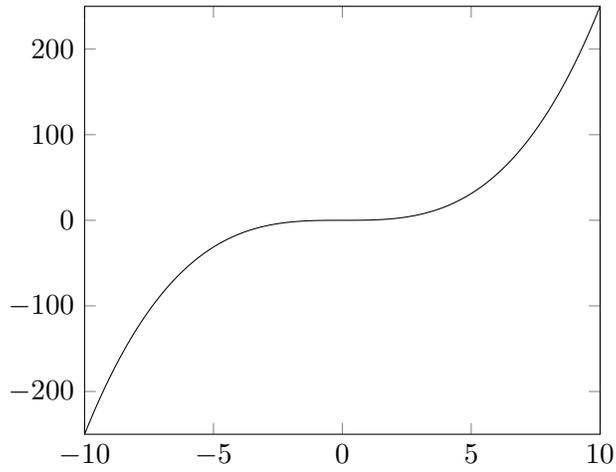
\vspace{2em}

\begin{tikzpicture}
\begin{axis}[enlargelimits=false]
\addplot [domain=-10:10, samples=101,unbounded coords=jump]{(x^3)/4)};
\end{axis}
\end{tikzpicture}

\vspace{2em}

The derivative of hyper-sinh for positive values can be expressed as:

\vspace{1em}
\begin{math}
{\cosh{\left(x\right)} \times \frac{1}{3}}
\end{math}
\hspace{31.5em}(3)
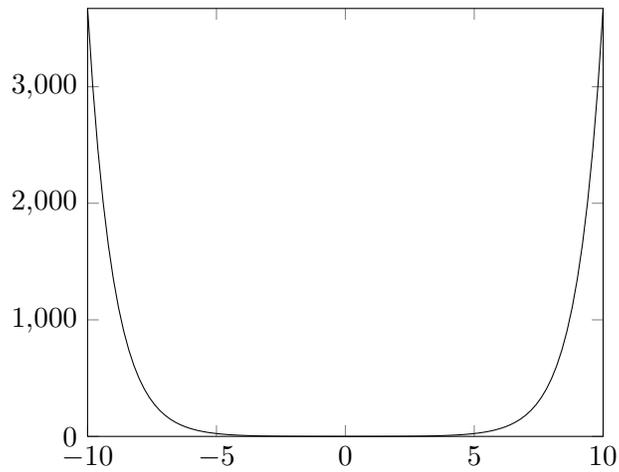
\vspace{2em}

\begin{tikzpicture}
\begin{axis}[enlargelimits=false]
\addplot [domain=-10:10, samples=101,unbounded coords=jump]{cosh(x)/3};
\end{axis}
\end{tikzpicture}

\vspace{2em}

The derivative of hyper-sinh for negative values and 0 (zero) can be expressed as:

\vspace{1em}
\begin{math}
{x^2 \times \frac{3}{4}}
\end{math}
\hspace{34.0em}(4)
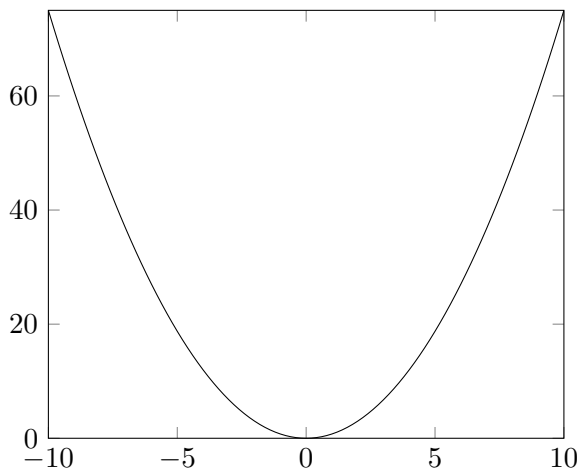
\vspace{2em}

\begin{tikzpicture}
\begin{axis}[enlargelimits=false]
\addplot [domain=-10:10, samples=101,unbounded coords=jump]{(x^2)*(3/4)};
\end{axis}
\end{tikzpicture}

\vspace{2em}

Listing 5 provides the snippet of code in Python that implements the proposed hyper-sinh function as an activation function and its derivative in TensorFlow~\citep{abadi2016tensorflow}.

\lstset{language=Python}
\lstset{frame=lines}
\lstset{caption={Using the hyper-sinh function as an activation function in TensorFlow~\citep{abadi2016tensorflow}.}}
\lstset{label={lst:code_direct_5}}
\lstset{basicstyle=\footnotesize}
\begin{lstlisting}

# Defining the hyper-sinh function

import numpy as np

def hyper_sinh(x):

  if x>0:
    x = 1/3*np.sinh(x)
    return x

  else:
    x = 1/4*(x**3)
    return x

# Vectorising the hyper-sinh function  
np_hyper_sinh = np.vectorize(hyper_sinh)

# Defining the derivative of the function hyper-sinh

def d_hyper_sinh(x):

  if x>0:
    x = 1/3*np.cosh(x)
    return x

  else:
    x = 3/4*(x**2)
    return x
          
np_d_hyper_sinh = np.vectorize(d_hyper_sinh)

# Defining the gradient function of the hyper-sinh

def hyper_sinh_grad(op, grad):
    x = op.inputs[0]
    n_gr = tf_d_hyper_sinh(x)
    return grad * n_gr

def py_func(func, inp, Tout, stateful=True, name=None, grad=None):
# Generating a unique name to avoid duplicates
    rnd_name = 'PyFuncGrad' + str(np.random.randint(0, 1E+2))
    tf.RegisterGradient(rnd_name)(grad)
    g = tf.get_default_graph()
    with g.gradient_override_map({"PyFunc": rnd_name}):
        return tf.py_func(func, inp, Tout, stateful=stateful, name=name)
        
np_hyper_sinh_32 = lambda x: np_hyper_sinh(x).astype(np.float32)
def tf_hyper_sinh(x,name=None):
    with tf.name_scope(name, "hyper_sinh", [x]) as name:
        y = py_func(np_hyper_sinh_32,   #forward pass function
                        [x],
                        [tf.float32],
                        name=name,
                         grad= hyper_sinh_grad) # The function that overrides gradient
        y[0].set_shape(x.get_shape())     # Specify input rank
        return y[0]
np_d_hyper_sinh_32 = lambda x: np_d_hyper_sinh(x).astype(np.float32)
def tf_d_hyper_sinh(x,name=None):
    with tf.name_scope(name, "d_hyper_sinh", [x]) as name:
        y = tf.py_func(np_d_hyper_sinh_32,
                        [x],
                        [tf.float32],
                        name=name,
                        stateful=False)
        return y[0]

\end{lstlisting}

Listing 6 provides the snippet of code in Python that implements the proposed hyper-sinh function in Keras~\citep{chollet2015keras}.

\lstset{language=Python}
\lstset{frame=lines}
\lstset{caption={Using the hyper-sinh function as an activation function in Keras~\citep{chollet2015keras}.}}
\lstset{label={lst:code_direct_6}}
\lstset{basicstyle=\footnotesize}
\begin{lstlisting}

from tensorflow.keras.layers import Layer

class hyper_sinh(Layer):

    def __init__(self):
        super(hyper_sinh,self).__init__()

    def build(self, input_shape):
        super().build(input_shape)

    def call(self, inputs,name=None):
        return tf_hyper_sinh(inputs,name=None)

    def get_config(self):
        base_config = super(hyper_sinh, self).get_config()
        return dict(list(base_config.items()))

    def compute_output_shape(self, input_shape):
        return input_shape

\end{lstlisting}

\vskip 0.15in

\subsection{Performance evaluation}

\vskip 0.15in

The accuracy of the FC-NN and CNN using different activation functions as described in sub-sections 2.2 and 2.3 on the data sets outlined in sub-section 2.1, was evaluated via the \emph{'accuracy\_score'} available in 'scikit-learn'~\citep{scikit-learn} from \emph{'sklearn.metrics'}. 
The reliability of such classifiers was assessed via the weighted average of the precision, recall and F1-score computed via the \emph{'classification\_report'}, also available in 'scikit-learn'~\citep{scikit-learn} from \emph{'sklearn.metrics'}.
\\
To understand what classification accuracy and reliability are, and how they can be evaluated, please refer to the following studies:~\citep{parisi2018decision},~\citep{parisi2018feature},~\citep{parisi2020novel},~\citep{parisi2020evolutionary}.


\section{Results}

\vskip 0.15in

Experimental results support the application of the proposed hyper-sinh activation function for both image and text classification tasks, as being accurate and reliable with the following classification performance:
\begin{itemize}
\item For shallow neural networks (FC-NN): 
\begin{itemize}
\item The \nth{2} highest accuracy on 2 out of 5 data sets evaluated (Tables 4 and 5 on text classification).
\item The \nth{2} highest reliability on 2 out of 4 data sets evaluated (Tables 4 and 5 on text classification).
\end{itemize}

\item For deep neural networks (CNN): 
\begin{itemize}
\item The best classification performance on 1 out of 5 data sets evaluated (Table 3).
\item The \nth{2} highest classification performance on 1 out of 5 data sets evaluated (Table 1).
\item The \nth{2} highest accuracy on 1 out of 5 data sets evaluated (Tables 1 on image classification).
\item The \nth{2} highest reliability on 1 out of 5 data sets evaluated (Table 1 on image classification).
\end{itemize}
\end{itemize}

\vskip 0.145in

\newpage


\textbf{Table 1.} Results on performance evaluation of baseline (non-optimised) three-layered Convolutional Neural Network (CNN) in Keras with different activation functions, including the proposed hyper-sinh function. The performance of such classifiers was evaluated on the ‘CIFAR-10’ data set available in Keras. 

\begin{table}[H]
\resizebox{\textwidth}{!}{%
\begin{tabular}{lllllll}
\textbf{Classifier} &
  \textbf{Activation function} &
  \textbf{Epochs} &
  \textbf{\begin{tabular}[c]{@{}c@{}}Testing accuracy \\ (0-1)\end{tabular}} &
  \textbf{\begin{tabular}[c]{@{}c@{}}Weighted precision \\ (0-1)\end{tabular}} &
  \textbf{\begin{tabular}[c]{@{}c@{}}Weighted recall \\ (0-1)\end{tabular}} &
  \textbf{\begin{tabular}[c]{@{}c@{}}Weighted F1-score\\ (0-1)\end{tabular}} \\ 
\textbf{CNN} & \begin{tabular}[c]{@{}c@{}}hyper-sinh\\ (this study)\end{tabular} & 10 & 0.70 & 0.70 & 0.70 & 0.69 \\
\textbf{CNN} & ReLU                                                              & 10 & 0.71 & 0.71 & 0.71 & 0.71 \\
\textbf{CNN} & sigmoid                                                           & 10 & 0.10 & 0.01 & 0.10 & 0.02 \\
\textbf{CNN} & tanh                                                              & 10 & 0.69 & 0.69 & 0.69 & 0.69 \\
\end{tabular}%
}
\end{table}


\textbf{Table 2.} Results on performance evaluation of baseline (non-optimised) Fully Connected Neural Network (FC-NN) with one hidden layer having 128 neurons in Keras with different activation functions, including the proposed hyper-sinh function. The performance of such classifiers was evaluated on the ‘Fashion-MNIST’ data set available in Keras. 

\begin{table}[H]
\resizebox{\textwidth}{!}{%
\begin{tabular}{lllllll}
\textbf{Classifier} &
  \textbf{Activation function} &
  \textbf{Epochs} &
  \textbf{\begin{tabular}[c]{@{}c@{}}Testing accuracy \\ (0-1)\end{tabular}} &
  \textbf{\begin{tabular}[c]{@{}c@{}}Weighted precision \\ (0-1)\end{tabular}} &
  \textbf{\begin{tabular}[c]{@{}c@{}}Weighted recall \\ (0-1)\end{tabular}} &
  \textbf{\begin{tabular}[c]{@{}c@{}}Weighted F1-score\\ (0-1)\end{tabular}} \\ 
\textbf{FC-NN} & \begin{tabular}[c]{@{}c@{}}hyper-sinh\\ (this study)\end{tabular} & 20 & 0.85 & 0.87 & 0.85 & 0.86 \\ 
\textbf{FC-NN} & ReLU                                                              & 20 & 0.88 & 0.88 & 0.88 & 0.88 \\ 
\textbf{FC-NN} & sigmoid                                                           & 20 & 0.89 & 0.89 & 0.89 & 0.89 \\ 
\textbf{FC-NN} & tanh                                                              & 20 & 0.88 & 0.89 & 0.88 & 0.88 \\ 
\end{tabular}%
}
\end{table}


\textbf{Table 3.} Results on performance evaluation of baseline (non-optimised) two-layered Convolutional Neural Network (CNN) in Keras with different activation functions, including the proposed hyper-sinh function. The performance of such classifiers was evaluated on the ‘MNIST’ data set available in Keras.

\begin{table}[H]
\resizebox{\textwidth}{!}{%
\begin{tabular}{lllllll}
\textbf{Classifier} &
  \textbf{Activation function} &
  \textbf{Epochs} &
  \textbf{\begin{tabular}[c]{@{}c@{}}Testing accuracy \\ (0-1)\end{tabular}} &
  \textbf{\begin{tabular}[c]{@{}c@{}}Weighted precision \\ (0-1)\end{tabular}} &
  \textbf{\begin{tabular}[c]{@{}c@{}}Weighted recall \\ (0-1)\end{tabular}} &
  \textbf{\begin{tabular}[c]{@{}c@{}}Weighted F1-score\\ (0-1)\end{tabular}} \\ 
\textbf{CNN} & \begin{tabular}[c]{@{}c@{}}hyper-sinh\\ (this study)\end{tabular} & 15 & 0.99 & 0.99 & 0.99 & 0.99 \\ 
\textbf{CNN} & ReLU                                                              & 15 & 0.99 & 0.99 & 0.99 & 0.99 \\ 
\textbf{CNN} & sigmoid                                                           & 15 & 0.98 & 0.98 & 0.98 & 0.98 \\ 
\textbf{CNN} & tanh                                                              & 15 & 0.99 & 0.99 & 0.99 & 0.99 \\ 
\end{tabular}%
}
\end{table}


\textbf{Table 4.} Results on performance evaluation of baseline (non-optimised) Fully Connected Neural Network (FC-NN) with one hidden layer having 512 neurons in Keras with different activation functions, including the proposed m-sinh function. The performance of such classifiers was evaluated on the ‘Reuters’ data set available in Keras.

\begin{table}[H]
\resizebox{\textwidth}{!}{%
\begin{tabular}{lllllll}
\textbf{Classifier} &
  \textbf{Activation function} &
  \textbf{Epochs} &
  \textbf{\begin{tabular}[c]{@{}c@{}}Testing accuracy \\ (0-1)\end{tabular}} &
  \textbf{\begin{tabular}[c]{@{}c@{}}Weighted precision \\ (0-1)\end{tabular}} &
  \textbf{\begin{tabular}[c]{@{}c@{}}Weighted recall \\ (0-1)\end{tabular}} &
  \textbf{\begin{tabular}[c]{@{}c@{}}Weighted F1-score\\ (0-1)\end{tabular}} \\
\textbf{FC-NN} & \begin{tabular}[c]{@{}c@{}}hyper-sinh\\ (this study)\end{tabular} & 3 & 0.80 & 0.79 & 0.80 & 0.79 \\ 
\textbf{FC-NN} & ReLU                                                              & 3 & 0.80 & 0.80 & 0.80 & 0.79 \\ 
\textbf{FC-NN} & sigmoid                                                           & 3 & 0.80 & 0.79 & 0.80 & 0.78 \\ 
\textbf{FC-NN} & tanh                                                              & 3 & 0.81 & 0.81 & 0.81 & 0.80 \\ 
\end{tabular}%
}
\end{table}

\newpage


\textbf{Table 5.} Results on performance evaluation of baseline (non-optimised) Fully Connected Neural Network (FC-NN) with one hidden layer having 512 neurons in Keras with different activation functions, including the proposed m-sinh function. The performance of such classifiers was evaluated on the ‘IMDB’ data set available in Keras.

\begin{table}[H]
\resizebox{\textwidth}{!}{%
\begin{tabular}{lllllll}
\textbf{Classifier} &
  \textbf{Activation function} &
  \textbf{Epochs} &
  \textbf{\begin{tabular}[c]{@{}c@{}}Testing accuracy \\ (0-1)\end{tabular}} &
  \textbf{\begin{tabular}[c]{@{}c@{}}Weighted precision \\ (0-1)\end{tabular}} &
  \textbf{\begin{tabular}[c]{@{}c@{}}Weighted recall \\ (0-1)\end{tabular}} &
  \textbf{\begin{tabular}[c]{@{}c@{}}Weighted F1-score\\ (0-1)\end{tabular}} \\ 
\textbf{FC-NN} & \begin{tabular}[c]{@{}c@{}}hyper-sinh\\ (this study)\end{tabular} & 3 & 0.86 & 0.87 & 0.86 & 0.86 \\ 
\textbf{FC-NN} & ReLU                                                              & 3 & 0.87 & 0.87 & 0.87 & 0.87 \\ 
\textbf{FC-NN} & sigmoid                                                           & 3 & 0.86 & 0.86 & 0.86 & 0.86 \\ 
\textbf{FC-NN} & tanh                                                              & 3 & 0.86 & 0.86 & 0.86 & 0.86 \\ 
\end{tabular}%
}
\end{table}

\newpage


\vskip 0.3in

\section{Discussion}

\vskip 0.15in

As demonstrated by the competitive results obtained on the 5 data sets evaluated, especially those in Tables 1 and 3 for the deep neural network CNN and Tables 4 and 5 for the shallow neural network FC-NN, the hyper-sinh is deemed a suitable activation function that scales from shallow to deep neural networks.
\\
In fact, its accuracy and reliability was high across both sets of benchmark image- and text-based data sets, as quantified via appropriate metrics in sub-section 2.4, and better than some gold standard functions, e.g., considering Table 1 with the accuracy and the F1-score of the CNN using hyper-sinh being 0.70 and 0.69 respectively on the CIFAR-10 image-based data set, as opposed to that of the same CNN but using sigmoid being 0.10 and 0.02 respectively. Moreover, its accuracy and reliability were comparable to the FC-NN using ReLU (accuracy = 0.80, F1-score = 0.79), with higher reliability than the same FC-NN when leveraging the sigmoid function on the 'Reuters' text-based data set (F1-score = 0.78). The proposed hyper-sinh also led to increased precision on the 'IMDB' text-based data set (precision = 0.87) as opposed to sigmoid and tanh (precision = 0.86), when using the same FC-NN as that leveraged to classify the 'Reuters' data set.
\\
Therefore, the hyper-sinh demonstrates that it is possible to extend the m-arcsinh to generalise across both shallow and deep neural networks for image and text classification tasks, and that the mathematical formulation of this extended function does not have to be complex at all.
As an accurate and reliable activation function, the hyper-sinh is thus deemed a new gold standard activation function for both shallow and deep neural networks, freely available in TensorFlow and Keras. 


\section{Conclusion}

hyper-sinh was proven an accurate and robust activation function for shallow and deep neural networks for image and text classification, thus being a new gold standard that scales well for FC-NN and CNN. 
Since it is made freely available, open source, on the Python, TensorFlow and Keras ecosystems, it adds to the selection of activation functions that both not-for-profit and for-profit organisations can have when tackling image and text classification tasks with data sets of various sizes. 
Importantly, the proposed algorithm, being accurate and reliable, and written in a high-level programming language (Python), can be leveraged as a part of ML-based pipelines in specific use cases, wherein high accuracy and reliability need to be achieved, such as in the healthcare sector (e.g., in counselling psychology), from small to large clinics with its suitability from shallow to deep neural networks.
Future work involves further improving this function to reduce its computational cost.


\acks{This research did not receive any specific grant from funding agencies in the public, commercial, or not-for-profit sectors.}


\newpage

\bibliography{ArXiv_paper_LP}

\end{document}